\documentclass[10pt,twocolumn,letterpaper]{article}

\usepackage{iccv}      

\usepackage[utf8]{inputenc} 
\usepackage[T1]{fontenc}    
\usepackage{graphicx}
\usepackage{amsmath}
\usepackage{amssymb}
\usepackage{epsfig}

\usepackage{booktabs}       
\usepackage{multirow}
\usepackage{colortbl}       
\usepackage{array}          
\usepackage{makecell} 
\usepackage{diagbox}
\usepackage{caption}
\usepackage{subcaption}     
\usepackage{wrapfig}        
\usepackage{float}          
\usepackage{pifont}         

\usepackage{algorithm}
\usepackage{listings}

\definecolor{iccvblue}{rgb}{0.21,0.49,0.74}
\usepackage[pagebackref,breaklinks,colorlinks,allcolors=iccvblue]{hyperref}
\usepackage[accsupp]{axessibility}  

\ifdefined\isarxiv
\usepackage{url}            
\fi

\def\paperID{14221} 
\def\confName{ICCV}
\def\confYear{2025}

\title{CVPT: Cross Visual Prompt Tuning}

\renewcommand{\thefootnote}{\dagger}
\author{
\normalsize Lingyun Huang$^{1}$, \quad
Jianxu Mao$^{1}$\thanks{Corresponding Authors.}, \quad
Junfei Yi$^{1}$\footnotemark[1], \quad
Ziming Tao$^{1}$, \quad
Yaonan Wang$^{1}$ \\
\normalsize $^1$Hunan University\\
{\tt\small h15200228261@gmail.com, \{maojianxu, yijunfei, taozimingphd, yaonan\}@hnu.edu.cn}
}

\begin{document}
\maketitle
\renewcommand{\thefootnote}{\arabic{footnote}}
\begin{abstract}
Parameter-Efficient Fine-Tuning (PEFT) has emerged to mitigate the computational demands of large-scale models. Within computer vision, adapter-based PEFT methods are often favored over prompt-based approaches like Visual Prompt Tuning (VPT) due to the latter's performance and efficiency limitations. Our analysis reveals that VPT's shortcomings stem from its prompt deployment strategy, which can distort the model's inherent self-attention mechanism. To address this, we propose Cross Visual Prompt Tuning (CVPT). CVPT introduces a cross-attention module to directly model interactions between prompts and image tokens. This design decouples the prompts from the input sequence, preserving the original self-attention integrity while enabling efficient feature integration. Furthermore, we employ a weight-sharing mechanism for cross-attention initialization, which enhances representative capability without a large parameter overhead. Extensive experiments across 25 datasets show that CVPT significantly outperforms VPT. For instance, on the VTAB-1K benchmark, CVPT achieves over 4\% higher average accuracy, rivaling leading adapter-based methods in both performance and efficiency. Our work confirms that prompt-based methods can achieve exceptional results in visual fine-tuning. The code is available at \url{https://github.com/Lingyun0419/CVPT}
\end{abstract}    
\section{Introduction}
Increasing the scale of the models is a common method to enhance the model's performance \cite{clip}\cite{bert}\cite{bart}\cite{Roberta}. In recent years, with the rapid development of computing devices, model sizes have significantly increased \cite{xlnet}\cite{clark2020electra}\cite{deberta}\cite{vitg}. For instance, the number of parameters in the GPT series developed by OpenAI has surged from 117 million to 1.8 trillion in just five years \cite{gpt-1}\cite{gpt-2}\cite{gpt3}. The rapidly increasing number of parameters will lead to the problem of immense computational overhead. Therefore, adapting those models to downstream tasks with the full-tuning method will incur enormous costs. To resolve this issue, the PEFT approach has been proposed \cite{lora}\cite{prompt-tuning}\cite{VP}\cite{vl-adapter}\cite{vit-adapter}. PEFT adapts those large-scale pre-trained models to downstream tasks in a more efficient way by fine-tuning a subset of the models that contain much fewer parameters. Two mainstream methods within PEFT are Adapter \cite{adapter} and Prompt \cite{prompt-tuning}. During the training process, the Adapter inserts adapters into each transformer block and tunes those adapters, while the Prompt inserts prompt tokens into the embedded tokens to update the prompt tokens. 

\begin{figure}[t]
\centering
    \vspace{0em}
    \includegraphics[width=0.4\textwidth]{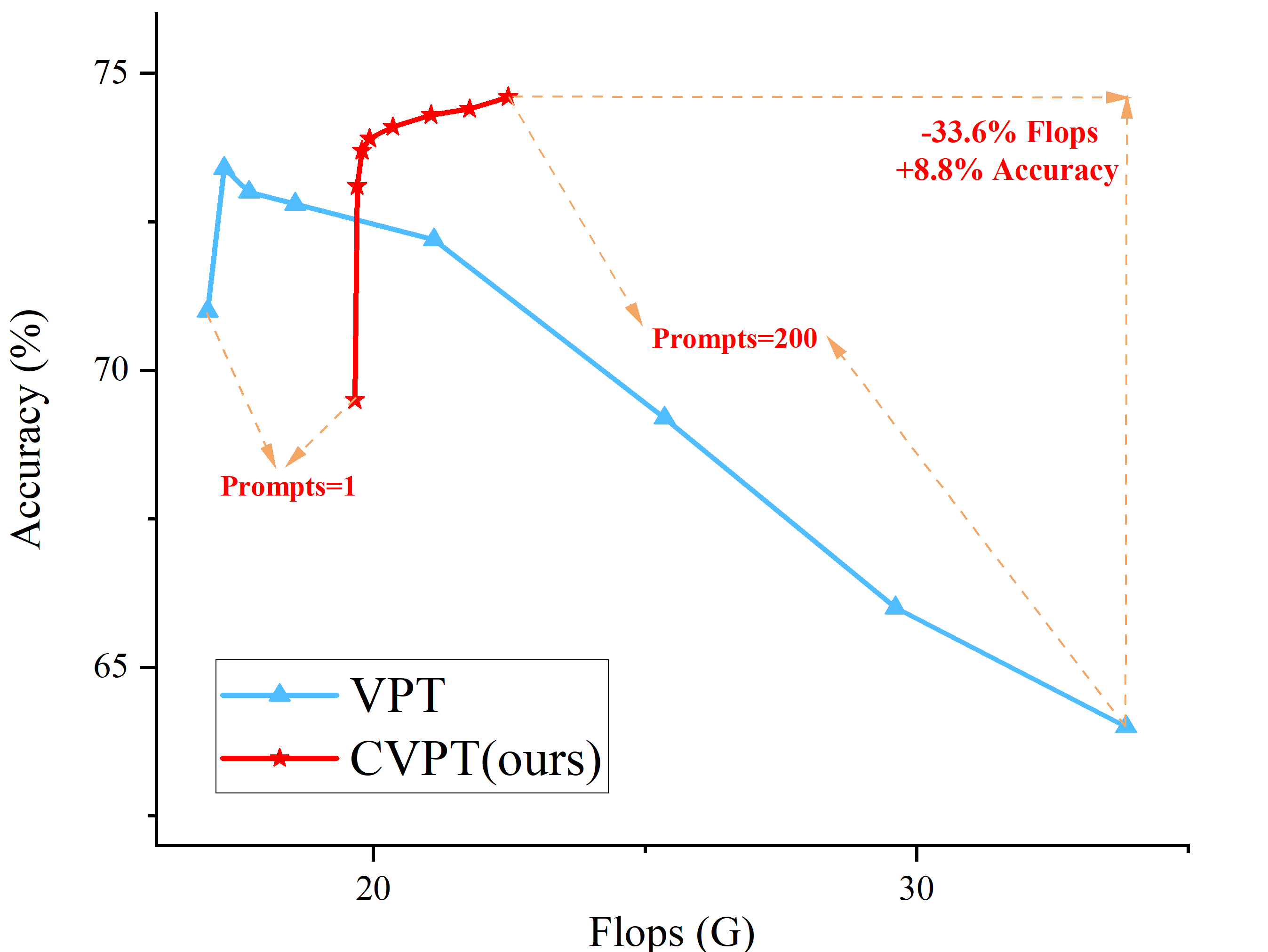} 
    \vspace{-0.5em}
    \caption{\textbf{Comparisons of performance and flops between VPT and our CVPT} with a pre-trained ViT-B/16 model on the VTAB-1k benchmark. We set the number of prompts to 1, 10, 20, 50, 100, 150, 200 respectively.}  
    \label{fig1}
    \vspace{-1em}
\end{figure}

VPT, a prompt-based method is first introduced by Jia \emph{et al.}~\cite{vpt} for visual fine-tuning tasks. Nevertheless, research on the adapter-based method is prominent due to its superior performance. Although some works have improved the performance of VPT \cite{dam}\cite{e2vpt}\cite{express}, it is still challenging to match the effectiveness to that of adapter-based methods. There appears to be a consensus that prompt-based methods underperform adapter-based methods in the visual domain. But is that really the case?

We conduct extensive experiments and analyses on VPT to uncover the reasons for its weaker performance compared to the Adapter. As a result of our investigation, we attribute this issue to the deployment of prompts in VPT. In VPT, prompts are concatenated with embedded tokens and processed together by the transformer blocks. Notably, this concatenation occurs along the token dimension, leading to the computational complexity of self-attention increasing quadratically with the number of prompts, which introduces computational redundancy. Moreover, the self-attention among embedded tokens is influenced by the prompt tokens, thereby distorting the original features \ref{sec3}. This implies that prompt-based methods are constrained to use only a limited number of prompts. However, due to computational inefficiencies and the limited learnable parameter count, \textbf{this smaller number of prompts lacks the flexibility to adapt to various downstream tasks}, resulting in a performance gap between VPT and Adapter methods. 

To address the issues above, it is necessary to alter the deployment of prompts to decouple prompts from the self-attention of embedded tokens. However, this also implies the need to find an alternative way to establish connections between embeddings and prompts, enabling prompts to participate in the fine-tuning process. With this in mind, we propose CVPT. We use cross-attention to capture the relationship between prompts and embedded tokens, incorporating the result as a residual term into the embedded tokens. This approach avoids the computational complexity of self-attention that is quadratically related to the number of prompts and allows prompts to focus on the embedded token to adapt to downstream tasks more efficiently. Additionally, by maintaining consistency in token dimensions throughout the computation process, the results of cross-attention can be directly summed with embedded tokens as residuals and do not introduce additional computational overhead for subsequent MLP. Furthermore, we share the weights of the self-attention layer with the cross-attention layer during loading checkpoints, keeping the cross-attention layer frozen alongside the self-attention layer, which eliminates the requirement for additional learned parameters for the cross-attention, and utilizes the encoded information in self-attention to help the fine-tuning of the model. In \ref{fig1}, we present the trends of accuracy and flops for CVPT and VPT with different numbers of prompts.

We validate the effectiveness of our method on 25 datasets, the results show that the CVPT achieves a significant improvement in performance and efficiency compared to the VPT. CVPT shows an average \textbf{5\%} improvement in accuracy on the 19 VTAB-1K datasets, \textbf{1\%} on the 5 FGVC datasets, and \textbf{3\%} on the ADE20K dataset. Additionally, benefiting from the design of CVPT, it achieves superior performance compared to other prompt-based methods, even with a limited number of prompts. If a large number of prompts is allowed, CVPT outperforms the SOTA methods on out-of-distribution datasets like ADE20K. Besides, although a large number of prompts are used, it does not introduce too much extra computational overhead compared.

Finally, we explore the impact of the deployment's position and the effectiveness of the weight-sharing mechanism. The improvement on the model can be fully illustrated by the experimental results above, indicating that prompt-based methods can also rival advanced adapter-based methods.

Overall, our contributions are as follows:

\begin{itemize}
\item We provide a detailed analysis of the application of VPT to visual tasks, and propose that its drawback can be attributed to \textbf{inefficiency and redundancy of computation} and \textbf{the destruction of self-attention}, which are caused by the deployment of prompts.
\item We propose CVPT, which introduces cross-attention and weight-sharing mechanisms, to decouple prompts from the self-attention of embedded tokens, which allows prompts to integrate visual features efficiently. This makes it possible to use a large number of prompts to adapt to downstream tasks, thereby improving both performance and efficiency.
\item We conducted experiments on 25 datasets with different downstream tasks. The results show that our approach significantly outperforms the original VPT and other prompt-based works in terms of performance and efficiency. It is also comparable to advanced adapter-based methods, demonstrating the usability of the prompt-based approach for visual fine-tuning.
\end{itemize}
\section{Related Work}

\textbf{PEFT. }In the era of CNN, making bigger and deeper models was an effective way to improve performance \cite{alex}\cite{resnet}\cite{resnext}. With the rise of transformers, this trend became even more popular. ChatGPT's introduction further cemented the community's goal to develop larger and more powerful models. However, limited by their scale, despite their powerful performance and generality, these large models are difficult to adapt to downstream tasks by using traditional paradigms (full-tuning). Consequently, NLP researchers first proposed PEFT methods. Their works demonstrate that fine-tuning just a small number of parameters in a large-scale pre-trained model can achieve nearly the same performance as full-tuning. Encouraged by the success in NLP, researchers began to apply PEFT to large-scale vision models on different visual tasks \cite{bayesian-prompt}\cite{aim}. After development in the past several years, the mainstream PEFT methods can be broadly categorized into adapter-based methods and Prompt-based methods. We mainly introduce prompt-based methods.

\noindent\textbf{Prompt. }Prompt was originally used in the field of NLP which is added to the input text for comprehension tasks. Lester \emph{et al.}~\cite{prompt-tuning} proposed treating the prompt as a continuous vector and fine-tuning the model by updating its gradients. Jia \emph{et al.}~\cite{vpt} introduced this concept to visual fine-tuning for the first time, naming it VPT. As shown in Fig.\ref{fig3}, the embedded tokens are spliced with the prompt tokens before entering each transformer block, allowing it to participate in every layer of the network within the transformer block. Before entering the next transformer block, the prompt tokens of the previous layer are discarded, and new prompt tokens are spliced with the embedded token again (VPT-Deep). This can be formulated as shown below:
\begin{align}
\label{eq:vpt}
    [\vec{x}_i, \underline{\hspace{0.3cm}}, \vec{E}_i] &= \textcolor[rgb]{0.1,0.8,0.9}{L_i}([\vec{x}_{i-1}, \textcolor[RGB]{202,12,22}{\vec{P}_{i-1}},\vec{E}_{i-1}]),
\end{align}
where the red and blue indicate learnable and frozen parameters, respectively. $P$ denotes a learnable d-dimensional vector, X is the CLS token, and E is the patched image. There are improved variants based on VPT, such as E2VPT \cite{e2vpt}, EXPRESS \cite{express} and DAM-VP \cite{dam}. In E2VPT, prompts are only combined with the key and value matrices in self-attention. However, these variants do not alter the deployment of prompts in VPT, leading to similar issues as those in VPT, and they are still unable to compete with advanced adapter-based methods.
\section{Analysis of Previous VPT}
The underperformance of VPT relative to adapter-based methods in the realm of PEFT has become a widely accepted notion. We aim to conduct an in-depth analysis of VPT to elucidate the root causes of its limitations. Specifically, employing a large number of prompts during fine-tuning can, in some cases, lead to significant performance improvements. In VPT-Deep, Jia \emph{et al.}~\cite{vpt} utilized 50+ prompts on 13 out of 24 datasets to achieve optimal performance, while in VPT-Shallow, 19 datasets benefited from the use of 50+ prompts. Subsequent works~\cite{express, dam} have leveraged even larger numbers of prompts to enhance performance, with APT~\cite{apt} employing over 1000 prompts. However, this increase in the number of prompts introduces the following challenges:
\label{sec3}
\subsection{Computational inefficiency and redundancy}

In VPT, prompts are concatenated with embedded tokens and transformer blocks process this combined input. Each token computes its attention score with every token when calculating multi-head self-attention. This results in the attention score matrix being divided into four parts: self-attention among the original tokens, self-attention among the prompts, attention from the original tokens to the prompts, and attention from the prompts to the original tokens. Since prompts are discarded at the end of each transformer block, we focus more on the embedded tokens that remain, specifically the self-attention matrix of the original tokens. We find that prompts do not directly affect the self-attention among the original tokens; instead, they can only indirectly fine-tune the embedded tokens by influencing other attention scores.

Besides, the computational complexity of self-attention is $n^2$, where $n$ is the number of embedded tokens. If $m$  represents the number of inserted prompt tokens, the computational complexity of self-attention in VPT can be expressed as $(n+m)^2$. This increases the computational overhead significantly, especially when using a larger number of prompt tokens. 

Additionally, since prompt tokens are discarded at the end of each transformer block, we found that those prompt tokens not only add computational overhead but also do not impact the results. Our experiments show that removing those prompt tokens after self-attention does not affect the results. And this phenomenon has been similarly noted in previous studies~\cite{apt}.

\subsection{Destruction of self-attention}
As we analyzed above, the attention score matrix is divided into four parts. Subsequently, the attention score matrix passes through a softmax layer. Along the channel dimension, the sum of the weights for each matrix is normalized to 1. This means that as the number of prompt tokens increases, the attention weights assigned to the self-attention among the embedded tokens are progressively weakened. Therefore, the ability to represent self-attention between embedded tokens will be weakened. Since the prompt token is eventually removed, this is equivalent to multiplying the self-attention result between the embedded tokens by a factor which less than one. To explore how large this effect is, we set the number of prompts to 1,5,20,50,100,150,196 respectively, and visualize the tensor after the softmax function, the results are shown in Fig.\ref{fig2} below.

\begin{figure}[h]
\centering
\vspace{-1em}
    \includegraphics[width=0.46\textwidth]{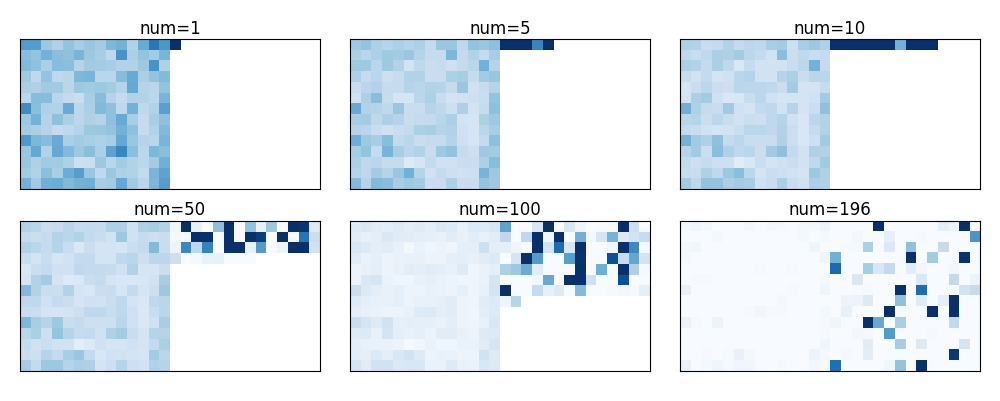}
    \vspace{-1em}
    \caption{
    \textbf{Self-attention weight obtained by prompt tokens and embedded tokens. } We visualize the self-attention of $cls_{token}$ and exclude itself to observe the attention of $cls_{token}$ to other tokens. Thus, we can understand the impact of different tokens on the attention scores for the $cls_{token}$. And the darker the color, the larger the weight. When giving 196 prompts, the attention weight obtained by prompts is over 80\%, which greatly influences the self-attention received by embedded tokens.}
    \label{fig2}
\vspace{-0.5em}
\end{figure}

As the number of prompts increases, the sum of the prompt's weight values exceeds 0.8, which is over 4 times that of embedded tokens, significantly disrupting the self-attention between the embedded tokens. This explains why VPT performance decreases substantially with a larger number of prompts.

\noindent\textbf{Previous solutions for excessive attention. }Several prior studies have also recognized the detrimental effects caused by this phenomenon of attention convergence. Xiao \emph{et al.}~\cite{attentionsink} posit that the performance collapse observed in large language models (LLMs) exceeding their maximum context length stems from attention convergence on non-informative initial tokens. This is attributed to the fact that, beyond the context length, initial tokens are excluded; however, because these tokens occupy a significant proportion of the softmax distribution, their exclusion induces a substantial shift in the distribution of attention scores, consequently leading to model failure. This observation parallels our own, as both are fundamentally rooted in the distributional issues of attention weights arising from the softmax function. They propose a mitigation strategy involving the introduction of additional learnable placeholder tokens to reduce the strong focus of other tokens on the initial tokens; this, in effect, acts as a form of smoothing for tokens that receive excessive attention.

Darcet \emph{et al.}~\cite{register} note that DINOv2 exhibits suboptimal performance when utilized for feature extraction. This is attributed to the presence of a subset of tokens possessing exceptionally high norms, thereby inducing an "attention-squeezing" effect on the remaining tokens. To mitigate this, they introduce several learnable REG tokens into the sequence, serving to smooth the scores of tokens exhibiting outlier values.

Previous prompt-based method also identified the issue of prompt tokens‘ excessive attention. In VFPT~\cite{vfpt}, they demonstrate that the attention weights acquired by the original tokens in their VFPT are significantly higher compared to those in VPT. This indicates that their approach also serves to smooth the attention distribution.

\subsection{Root causes of the limitations}
We observe that excessive attention allocated to prompt tokens is a recurring issue in prior works. Approaches that increase the attention weights of original tokens or decrease the attention weights of prompt tokens might appear to be viable solutions. However, these methods still face the same fundamental challenges as the number of inserted prompt tokens increases. Furthermore, they do not resolve the issue of the additional computational burden, which scales quadratically with the number of prompts.

We re-examine these approaches and find that, in the scenarios described in~\cite{attentionsink, register}, the tokens exhibiting anomalous attention originate from the original sequence. As intrinsic components of the original sequence, these tokens carry substantial semantic information, playing a crucial role in contextual understanding for natural language processing tasks or feature recognition in visual tasks. However, in VPT, the anomalous token - prompts are artificially introduced. They do not inherently possess semantic meaning, serving solely as an indirect tuning factor. Their importance is thus not comparable to that of the outlier tokens observed in the aforementioned scenarios, removing them does not disrupt the network's inherent capabilities. Based on this, we think that decoupling prompts from self-attention can fundamentally address the negative impact of prompts on self-attention.

\section{Method}
Having attributed the root cause of VPT's limitations to the involvement of prompts in the self-attention process, we recognize that decoupling prompts from self-attention necessitates a new approach to establish a connection between prompt tokens and original tokens, thereby enabling their participation in fine-tuning. Consequently, the core challenge transforms into \textbf{how to capture the semantic relationship between two distinct sequences.}

Inspired by the works in NLP, it occurred to us to use cross-attention to establish the connection between prompt and original tokens. Therefore, we introduce CVPT (Cross Visual Prompt Tuning). Next, we review cross-attention.

\begin{figure*}[t]
\centering
\vspace{-1.5em}
    \includegraphics[width=0.85\textwidth]{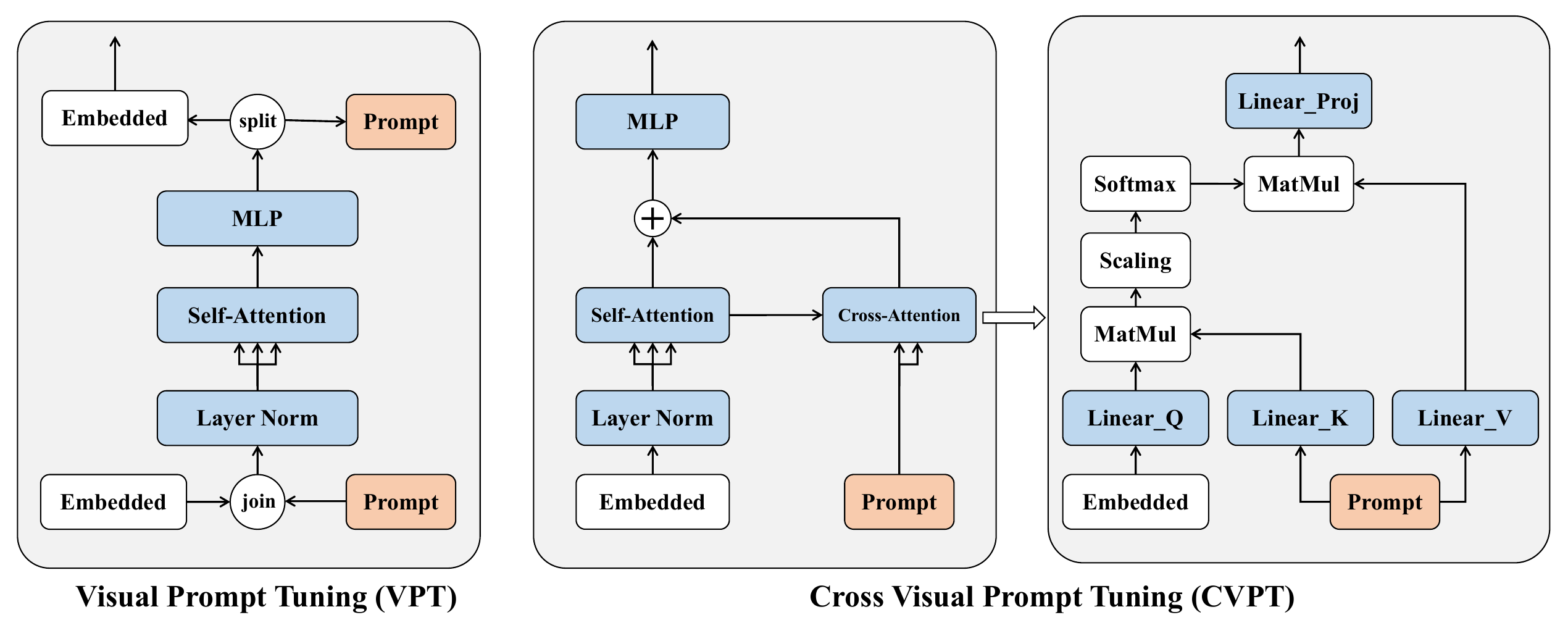}
        
    \caption{
    \textbf{Structure comparison of VPT and CVPT. }In which blue represents frozen parameters and orange represents learnable. 
    }
    \label{fig3}
\vspace{-1em}
\end{figure*}

\subsection{Cross-Attention}
Unlike self-attention \cite{vaswani2017attention}, which computes the relationship between each element in the input sequence, cross-attention computes attention on two different sequences to process the semantic relationship between them \cite{crossvit}. For example, in translation tasks, cross-attention is used to compute the attention weights between the source language sentence and the target language sentence. In our method, we introduce cross-attention to handle the semantic relationship between embedded tokens and prompt tokens, guiding the fine-tuning of the model. Specifically, the input of cross-attention consists of two parts: $X_1$ and $X_2$, in which $X_1  \in \mathbb{R}^{n \times d_1}$ and $X_2  \in \mathbb{R}^{m \times d_2}$. And $X_1$ serves as the query set and $X_2$ serves as the key-value set. We set $Q = X_1W^Q$ and $K = V = X_2W^K$, and then the cross-attention can be expressed as follows:
\begin{align}
\label{eq:CA}
CrossAttention(X_1,X_2) = Softmax\left ( \frac{Q\cdot K}{\sqrt{d_{k} } }  \right ) V.
\end{align}
In which $W^Q \in \mathbb{R}^{d_1 \times d_k}$ and $W^K \in \mathbb{R}^{d_2 \times d_k}$ are learned projection matrix, $d_k$ is the dimension of value-key set. In our methods, $d_1=d_2=d_k$. And the shape of output is $n \times d_k$, which is consistent with $X_1$.

\subsection{Cross Visual Prompt Tuning}
We propose CVPT, which redesigns the deployment of prompts to better adapt visual tasks. Our approach, as illustrated in Fig.\ref{fig3}, follows the VPT, the main parameters of the network remain frozen, and only the final classification layer and the prompt are trainable. The key difference is that we allow the prompt token to perform cross-attention with the embedded tokens and the result of cross-attention is added with the embedded tokens as residuals. This operation allows us to decouple the self-attention between prompts and embedded tokens, and more efficiently reestablish the connection between the prompts and embedded tokens. We demonstrate how significant this improvement is in Sec.\ref{sec5.2}. Specifically, for any input $x_i$ of a transformer block, the forward flow can be represented as follows:
\begin{align}
\label{eq:CVPT}
X_1=X_i+\textcolor[rgb] {0.1,0.8,0.9}{SA}(\textcolor[rgb] {0.1,0.8,0.9}{LN_1}(X_i)), \\
X_2=X_1+\textcolor[rgb] {0.1,0.8,0.9}{CA}(X_1, \textcolor[RGB]{202,12,22}{Prompt}), \\
X_{out}=X_2+\textcolor[rgb] {0.1,0.8,0.9}{MLP}(\textcolor[rgb] {0.1,0.8,0.9}{LN_2}(X_2)),
\end{align}

where blue denotes frozen parameters and red denotes trainable parameters, SA denotes self-attention, CA denotes cross-attention, and LN denotes layer normalization. 

In CVPT, we only introduce linear computational overhead associated with the number of prompt tokens. It allows CVPT to use a large number of prompt tokens to improve its performance by introducing an acceptable overhead. Furthermore, CVPT preserves the original procedure of self-attention, keeping the complete representation ability of embedded tokens. We demonstrate the improvement over VPT in terms of performance and efficiency in Sec.\ref{sec5.3}. Finally, we set embedded tokens as query sets and prompt tokens as key-value sets, so that the consistency of the dimension can be maintained, allowing the result of cross-attention to be directly summed with the input as a residual term.

\subsection{Weight-sharing mechanism}
The utilization of cross-attention, which requires a large number of learnable parameters, leads to a major challenge in computational overhead. Therefore, if the parameters of them are tunable, the computational overhead of CVPT will even rival those using full-tuning. Therefore, we introduce the weight-sharing mechanism. Due to the structure of cross-attention equal to that of self-attention, we consider that the weight of self-attention is also instructive for the fine-tuning of cross-attention. Thus, we initialize the weight of cross-attention with the parameters of self-attention when loading checkpoints. It avoids the introduction of a huge number of learnable parameters in cross-attention and keeps the efficiency of our CVPT. We explore the impact of weight-sharing in Sec.\ref{sec5.4} and demonstrate its effectiveness and efficiency.

\begin{table*}[t]
        \caption{\textbf{Performance comparisons on the VTAB-1k benchmark with ViT-B/16 models pre-trained on ImageNet-21K.}}
	\centering
        \vspace{-1em}
	\scalebox{0.68}{
        \begin{tabular}{ccc|ccccccc|cccc|cccccccc}
		\toprule
		& & &\multicolumn{7}{|c}{\textbf{Natural}} & \multicolumn{4}{|c}{\textbf{Specialized}} &  \multicolumn{8}{|c}{\textbf{Structured}} \\ \midrule
			Method& \rotatebox{90}{Params. (M)} & \rotatebox{90}{Avg. Acc.} & \rotatebox{90}{CIFAR-100} & \rotatebox{90}{Caltech101} & \rotatebox{90}{DTD} & \rotatebox{90}{Flowers102} & \rotatebox{90}{Pets} & \rotatebox{90}{SVHN}  & \rotatebox{90}{Sun397} & \rotatebox{90}{Patch Camelyon~} & \rotatebox{90}{EuroSAT}   & \rotatebox{90}{Resisc45}  & \rotatebox{90}{Retinopathy} & \rotatebox{90}{Clevr/count} & \rotatebox{90}{Clevr/distance}  & \rotatebox{90}{DMLab} & \rotatebox{90}{KITTI/distance~}  & \rotatebox{90}{dSprites/loc} & \rotatebox{90}{dSprites/ori}   & \rotatebox{90}{SmallNORB/azi~}  & \rotatebox{90}{SmallNORB/ele~}     \\
			\midrule
			Full-tuning&85.8&68.9&68.9&87.7&64.3&97.2&86.9&87.4&38.8&79.7&95.7&84.2&73.9&56.3&58.6&41.7&65.5&57.5&46.7&25.7&29.1\\
    Linear-probing~\cite{mae}&\bf0&57.6&63.4&85.0&63.2&97.0&86.3&36.6&51.0&78.5&87.5&68.6&74.0&34.3&30.6&33.2&55.4&12.5&20.0&9.6&19.2\\
    Bias~\cite{bitfit}&0.10&65.2&72.8&87.0&59.2&97.5&85.3&59.9&51.4&78.7&91.6&72.9&69.8&61.5&55.6&32.4&55.9&66.6&40.0&15.7&25.1\\
    Adapter~\cite{adapter}&0.15&73.9&69.2&90.1&68.0&98.8&89.9&82.8&54.3&84.0&94.9&81.9&75.5&80.9&65.3&48.6&78.3&74.8&48.5&29.9&41.6\\
    AdaptFormer~\cite{adaptformer}&0.15&74.7&70.8&91.2&70.5&99.1&90.9&86.6&54.8&83.0&95.8&84.4&\bf76.3&81.9&64.3&49.3&80.3&76.3&45.7&31.7&41.1\\
    LoRA~\cite{lora}&0.29&74.5&67.1&91.4&69.4&98.8&90.4&85.3&54.0&84.9&95.3&84.4&73.6&82.9&69.2&49.8&78.5&75.7&47.1&31.0&\bf44.4\\
    RepAdapter~\cite{repAdapter}&0.23&76.1&72.4&91.6&71.0&99.2&91.4&90.7&55.1&85.3&95.9&84.6&75.9&82.3&68.0&50.4&79.9&80.4&49.2&\bf38.6&41.0\\

    SPT-LoRA~\cite{spt}&0.48&76.4&73.5&\bf93.3&72.5&99.3&91.5&87.9&55.5&85.7&96.2&85.9&75.9&\bf84.4&67.6&52.5&\bf82.0&81.0&51.1&30.2&41.3\\

RLRR~\cite{rlrr}&0.33&76.8&\bf75.6& 92.4& 72.9& 99.3 &91.5 &89.8& \bf57.0 &\bf86.8& 95.2& 85.3&75.9 & 79.7& 64.2 &\bf53.9 &82.1 &\bf83.9 &\bf53.7& 33.4 &43.6 \\

DM-LoRA~\cite{dmlora}&0.29 &\bf77.0&74.0& 90.7 &\bf73.9 &\bf99.3 &\bf92.2 &\bf91.1& 56.4& 85.6 &\bf96.5 &\bf87.0 &76.1 &83.5&\bf69.9&52.0& 81.6 &80.2 &50.2& 36.1 &43.1 \\

    \midrule[0.5pt]
    VPT-shallow&\bf0.06&67.8&77.7&86.9&62.6&97.5&87.3&74.5&51.2&78.2&92.0&75.6&72.9&50.5&58.6&40.5&67.1&68.7&36.1&20.2&34.1\\
    VPT-Deep~\cite{vpt}&0.53&72.0&\bf78.8&90.8&65.8&98.0&88.3&78.1&49.6&81.8&96.1&83.4&68.4&68.5&60.0&46.5&72.8&73.6&47.9&32.9&37.8\\
    EXPRESS~\cite{express}&0.98&72.9&78.0&89.6&68.8&98.7&88.9&{89.1}&51.9&{84.8}&96.2&80.9&74.2&66.5&60.4&46.5&77.6&78.0&49.5&26.1&35.3\\
    DAM-VP~\cite{dam}&2.52&73.1&-&-&-&-&-&-&-&-&-&-&-&-&-&-&-&-&-&-&-\\
    $E^2$VPT~\cite{e2vpt}&0.27&73.9&78.6&89.4&67.8&98.2&88.5&85.3&52.3&\bf87.8&96.1&84.8&73.6&71.7&61.2&47.9&75.8&80.8&48.1&31.7&41.9\\

VAPT~\cite{vapt}&0.25&75.3&80.8&91.9&69.7&98.8&89.2&86.7&52.9&84.4&96.5&85.1&74.5 &74.8&63.6&50.0&77.2&86.1&48.3&33.8&40.9\\

VFPT~\cite{vfpt}&0.48&75.5&80.7 & 91.4 & 69.4 &\bf99.3& 90.3 &85.6&52.7&83.5&96.5&84.4&75.4&75.8&63.2&48.3&79.3&81.5&56.0&34.1&43.4\\

S$A^2$VP~\cite{sa2vp}&0.41&75.8&73.0&\bf91.9&70.5&99.1&90.8&84.7&\bf56.8&86.0&95.9&85.8&75.2&76.6&61.8&\bf50.8&79.9&\bf84.5&52.8&\bf34.7&45.3\\

   \rowcolor[RGB]{230, 230, 230}CVPT&0.39&\bf77.2&\bf81.5&91.5&\bf74.0&99.2&\bf91.4&\bf90.7&54.5&85.8&\bf96.5&\bf87.6&\bf75.8&\bf79.0&\bf67.2&50.6&\bf82.7&81.5&\bf53.0&34.4&\bf45.3\\
			\bottomrule
   \end{tabular}}
   \label{tab5-1}
    \vspace{-1em}
\end{table*}
\section{Experiment}
\subsection{Experimental settings}
\textbf{Datasets.} We evaluate our CVPT on both image classification and semantic segmentation tasks to verify its effectiveness. The specific datasets involved in our work are presented in the following.
\begin{itemize}
\item \textbf{VTAB-1K. }VTAB-1K comprises 19 datasets from different domains, classified into three main categories: the Natural group (natural images captured by standard cameras) \cite{cifar}\cite{svhn}\cite{caltech}\cite{pets}, the Specialized group (professional images captured by specialized equipment, such as medical and remote sensing images) \cite{patchcamelyon}\cite{erousat}, and the Structured group (synthetic images from artificial environments). Each task contains only 1,000 training samples \cite{clevr}\cite{kitti}\cite{matthey2017dsprites}. This is a primary metric for evaluating PEFT's performance.

\item\textbf{FGVC. }FGVC consists of five fine-grained visual classification benchmarks, including  CUB-200-2011 \cite{CUB}, NABirds \cite{birds}, Oxford Flowers \cite{flower}, Stanford-Dogs \cite{dogs} and Stanford-Cars \cite{cars}. Unlike VTAB-1K, the datasets in FGVC benchmarks are complete.

\item\textbf{ADE20K. }ADE20K \cite{ade20k} contains more than 25,000 images and is primarily used for scene perception, parsing, segmentation, multi-object recognition, and semantic understanding. This adaptation is challenging due to the huge gap between the objectives of pretraining and downstream tasks. 
\end{itemize}

\noindent\textbf{Baseline.} We primarily use CVPT to compare with the following methods: (1) Full-tuning and Linear Probing, (2) Adapter and its variants, and (3) VPT and its variants.

\noindent\textbf{Training.} We use the ViT-Base-16 model as our main model and AdamW as our optimizer. The other settings and training strategies follow those used in VPT. We set the number of prompts from [1, 5, 10, 20, 50, 100, 200] for VTAB-1K (consistent with VPT). Besides, we use a single NVIDIA 3090 on VTAB-1K and FGVC benchmark and use NVIDIA 3090 $\times$ 8 on ADE20k.

\subsection{Comparison with other PEFT methods}
\label{sec5.2}

\noindent\textbf{VTAB-1K. }We compared CVPT with other baselines on the VTAB-1K benchmark. The experimental results are shown in Table.\ref{tab5-1}, where we report the top-1 accuracy of these methods. In the table, we divide the prompt-based methods into one group and the other methods into another group. The bold values in each group represent the best accuracy.

We first compare our method with other prompt-based methods. The results of our experiments show that our method achieved the best performance among prompt-based methods in 12 out of 19 datasets, significantly outperforming VPT and other VPT-based methods. Notably, CVPT achieves the highest accuracy in 5 out of 8 datasets within the structured group, indicating that the addition of cross-attention significantly improves the adaptation of prompts. Therefore, CVPT performs better in those out-of-distribution (OOD) datasets. CVPT requires fewer parameters than the latest methods like VFPT~\cite{vfpt} and S$A^2$VP~\cite{sa2vp}, which also perform well.

When considering all PEFT methods, we find that on a small dataset like VTAB-1K, almost all mainstream PEFT methods outperformed full-tuning in terms of performance. This suggests that correctly selecting the parameters to fine-tune is crucial. Our CVPT, shows an impressive performance, over 0.2 in accuracy than DM-LoRA~\cite{dmlora} and 0.4 than RLRR~\cite{rlrr}, outperforming the other PEFT methods in performance in VTAB-1K. Notably, given the previously observed limitations of prompt-based methods in terms of performance and efficiency, our CVPT deeply explores the potential of prompt-based methods and demonstrates that prompt-based approaches can be competitive with state-of-the-art adapter-based methods.

\noindent\textbf{FGVC. }Performance on VTAB-1K alone is not enough to prove the superiority of CVPT. Therefore, we introduce the experimental results of CVPT on FGVC to explore its performance on a complete dataset of a certain scale. The results are shown in Table.\ref{tab5-2} below:

\begin{table}[h]
	\centering
    \vspace{-0.5em}
         \caption{\textbf{Performance comparisons on five FGVC datasets with ViT-B/16 models pre-trained on ImageNet-21K.}}
        \vspace{-1em}
	\setlength{\tabcolsep}{4pt}
	\scalebox{0.58}{
 \begin{tabular}{c|c|c|c|c|c|c|c}
			\toprule
			\diagbox{Method}{datasets}&\makecell[c]{~CUB-200~ \\ -2011} & ~NABirds~ & \makecell[c]{~Oxford~ \\ Flowers}  & \makecell[c]{~Stanford~ \\ Dogs}  & \makecell[c]{~Stanford~ \\ Cars}  & \makecell[c]{~~Avg.~~\\~~Acc.~~} & \makecell[c]{~Params. \\ (M)}   \tabularnewline \midrule
			Full fine-tuning & 87.3 & 82.7 & 98.8 & 89.4 & 84.5 & 88.5 & 86.0   \tabularnewline 
			Linear probing~\cite{mae} & 85.3 & 75.9 & 97.9 & 86.2 &  51.3 & 79.3 &  \bf0.18 \tabularnewline  
			Adapter ~\cite{adapter} & 87.1 & 84.3 & 98.5 & 89.8 & 68.6 & 85.7   &  0.41 \tabularnewline 
               AdaptFormer ~\cite{adaptformer} & 84.7 & 75.2 & 97.9 & 84.7 & 83.1 & 85.1   &  0.37 \tabularnewline 
LoRA~\cite{lora}&88.3&85.6&99.2&91.0&83.2&89.5&0.44\tabularnewline
SPT-LoRA~\cite{spt}& 88.6 & 83.4&99.5 &91.4 &\bf 87.3  &  90.1 & 0.48 \tabularnewline
RLRR~\cite{rlrr}&89.3&84.7&99.5&\bf92.0&87.0&90.4&0.47\tabularnewline
DMLoRA~\cite{dmlora}&\bf89.8&\bf86.6&\bf99.5&91.8&85.7&\bf90.7&0.47\tabularnewline

            \midrule[0.5pt]
			VPT-Shallow~\cite{vpt}  & 86.7 & 78.8 & 98.4 & 90.7 & 68.7 & 84.6 &\bf0.25 \tabularnewline 
			VPT-Deep~\cite{vpt}  & 88.5 & 84.2 & 99.0 & 90.2 & 83.6 & 89.1 &  0.85 \tabularnewline 
            DAM-VP~\cite{dam}  & 87.5 & 82.1 & 99.2 & \bf92.3 & - & - &  - \tabularnewline 
            EXPRESS~\cite{express}  & 88.3 & - & 99.0 & 90.0 & 80.5 & - & - \tabularnewline 
               $E^2$VPT~\cite{e2vpt}  & 88.5 & 84.2 & 99.0 & 90.2 & 83.6 & 89.2 &  0.45 \tabularnewline 
VAPT~\cite{vapt}&89.7&84.6&99.1&91.7&82.8&89.6&0.67\tabularnewline
VFPT~\cite{vfpt}&88.7&84.5&99.1&90.4&83.6&89.2&0.85\tabularnewline
S$A^2$VP~\cite{sa2vp}&89.1&85.8&99.3&92.1&84.1&90.1&0.85\tabularnewline

           \rowcolor[RGB]{230, 230, 230}CVPT   & \bf89.9 &\bf 86.5&\bf99.3 & 91.7 &\bf85.2  &  \bf90.5 & 0.77 \tabularnewline
			\bottomrule
		\end{tabular}
	\label{tab5-2}
    \vspace{-1em}
    }
\end{table}

Similar to the results on VTAB-1K, our approach substantially outperforms other prompt-based methods and achieves the best results in 4 out of 5 datasets in FGVC. It is only behind 0.2\% than DMLoRA~\cite{dmlora} and outperforms other methods. This demonstrates CVPT's generalization and adaptability to the increasing scale of data in the future.

\noindent\textbf{ADE20K. }Finally, we apply CVPT to SETR\cite{setr} on the ADE20K dataset to explore its performance on semantic segmentation tasks. The results are shown in Table.\ref{tab5-3}.

\begin{table}[h]
\vspace{-0.5em}
\caption{\textbf{Results of ADE20K datasets with ViT-L.} "mIoU-SS" and "mIoU-Ms" denote single-scale and multi-scale, respectively.}
\vspace{-1em}
\centering
\scalebox{0.95}{
\begin{tabular}{c|c|c|c}
\toprule
Methods&Params(M)&mIoU-SS&mIoU-Ms\\ 
\midrule[0.5pt]
Full-tuning&318.3&48.31&50.07\\ 
Linear probing&13.18&35.12&37.46\\ 
\midrule[0.5pt]
Bias~\cite{bitfit}&13.46&43.40&45.33\\
VPT~\cite{vpt}&13.43&42.11&44.06\\
VPT+Bias~\cite{vpt}&15.79&44.04&45.63\\
RepAdapter~\cite{repAdapter}&13.82&44.44&46.71\\
SPT-LoRA~\cite{spt}&14.60&45.40&47.50\\
\rowcolor[RGB]{230, 230, 230}CVPT(P=10)&\bf13.43&43.78&45.85\\
\rowcolor[RGB]{230, 230, 230}CVPT(P=200)&18.00&\bf45.66&\bf47.92\\
\bottomrule
\end{tabular}}
\label{tab5-3}
\vspace{-0.5em}
\end{table}  

This task is quite challenging because of the huge distribution gap between pre-training datasets and downstream tasks. In this situation, our CVPT shows a 1.7\% enhancement of "mIoU-SS" over the VPT with the same number of prompts. If we use 200 prompts for fine-tuning, CVPT represents a significant improvement over the other PEFT methods. This fully demonstrates the adaptation of CVPT to OOD datasets. Besides, due to our optimization of the deployment, even though the number of learnable parameters increases by 4 million, our memory usage and training time increase by less than 20\% compared to linear probing and less than 10\% compared to it when using 10 prompts during training.

\subsection{Comparison with VPT}
\label{sec5.3}
\textbf{Performance improvement. }To investigate how much improvement CVPT makes and the effect of the number of prompts on performance, we use different numbers of prompt tokens and conduct experiments on VTAB-1K using VPT and CVPT, respectively. The results are shown in the following Table.\ref{tab5-4}:

\begin{table}[h]
    \centering
    \vspace{-0.5em}
    \caption{\textbf{Performance comparisons With VPT and CVPT on VTAB-1K benchmark of different number of prompt tokens.}}
    \label{tab5-4}
    \vspace{-1em}
    \scalebox{0.71}{ 
        \begin{tabular}{c|c|c|c|c|c|c|c|c}
            \toprule
            \diagbox{Method}{Number} & 1 & 5 & 10 & 20 & 50 & 100 & 150 & 200 \\
            \midrule[0.5pt]
            VPT & \bfseries 71.0 & 73.0 & 73.0 & 72.8 & 72.2 & 69.2 & 66.0 & 64.0 \\ 
            \rowcolor[RGB]{230, 230, 230} CVPT & 69.5 & \bfseries 73.5 & \bfseries 74.0 & \bfseries 74.1 & \bfseries 74.3 & \bfseries 74.5 & \bfseries 74.6 & \bfseries 74.8 \\
            \bottomrule
        \end{tabular}
    }
    \vspace{-1em}
\end{table}

These results show that our CVPT achieves better performance in almost every case except the number of prompts equals 1. As we analyzed in Sec.\ref{sec3},  due to the destruction of self-attention between embedded tokens, when given a larger number of prompt tokens, VPT shows significant performance degradation or even crashes. In contrast, our CVPT avoids suffering from these problems. Additionally, its performance improves as the number of prompt tokens increases. We selected CIFAR\cite{cifar}, DTD, and SUN397 to investigate the reasons for the performance degradation in VPT, and the results are shown in Table.\ref{tab5-5}.

\begin{table}[h]
\centering
\vspace{-0.5em}
\caption{\textbf{Performance comparisons With VPT and CVPT on CIFAR, DTD, and SUN397,} whose feature distribution is similar to pretraining dataset.}
\vspace{-1em}
\scalebox{0.71}{ 
    \begin{tabular}{c|c|c|c|c|c|c|c}
    \toprule
    Dataset&\diagbox{Method}{Number} & 1 & 10 & 20 & 50 & 100 & 200 \\
    \midrule[0.5pt]
    \multirow{2}{*}{CIFAR} & VPT& 65.2 & 64.9 & 63.6 & 60.3 & 57.5 & 35.7\\
    &\cellcolor[RGB]{230, 230, 230}CVPT&\cellcolor[RGB]{230, 230, 230}\bfseries70.2&\cellcolor[RGB]{230, 230, 230}\bfseries72.4&\cellcolor[RGB]{230, 230, 230}\bfseries72.0&\cellcolor[RGB]{230, 230, 230}\bfseries72.6&\cellcolor[RGB]{230, 230, 230}\bfseries71.9&\cellcolor[RGB]{230, 230, 230}\bfseries72.1\\
    \midrule
    \multirow{2}{*}{DTD} & VPT& 68.8 & 66.1 & 65.9 & 63.4 & 61.9 & 59.5 \\
    & \cellcolor[RGB]{230, 230, 230}CVPT & \cellcolor[RGB]{230, 230, 230}\bfseries70.7 & \cellcolor[RGB]{230, 230, 230}\bfseries72.4 & \cellcolor[RGB]{230, 230, 230}\bfseries72.0 & \cellcolor[RGB]{230, 230, 230}\bfseries72.6 & \cellcolor[RGB]{230, 230, 230}\bfseries71.9 & \cellcolor[RGB]{230, 230, 230}\bfseries72.1 \\

    \midrule
    \multirow{2}{*}{SUN397} & VPT& \bfseries52.6 & 47.6 & 46.8 & 43.6 & 34.4 & 27.5 \\
    &\cellcolor[RGB]{230, 230, 230}CVPT&\cellcolor[RGB]{230, 230, 230}52.3&\cellcolor[RGB]{230, 230, 230}\bfseries54.4&\cellcolor[RGB]{230, 230, 230}\bfseries54.2&\cellcolor[RGB]{230, 230, 230}\bfseries54.1&\cellcolor[RGB]{230, 230, 230}\bfseries53.9 &\cellcolor[RGB]{230, 230, 230}\bfseries 54.0\\
    \bottomrule
    \end{tabular}}
\label{tab5-5}
\vspace{-0.5em}
\end{table}

Han \emph{et al.}~\cite{facing} think pretrained parameters play a pivotal role in capturing general features, while the added learnable parameters are important for potentially encoding task information in the context of transfer learning. This indicates that the destruction of self-attention between embedded tokens weakens the ability to recognize general features, and significantly leads to performance collapse on datasets with feature distributions similar to those of the pretraining data. Meanwhile, it can be observed that the performance of our CVPT does not fluctuate significantly with variations in the number of prompts. This demonstrates the advantage of preserving the complete self-attention.

\begin{figure*}[t]
\centering
    \vspace{-1em}
    \includegraphics[width=0.97\textwidth]{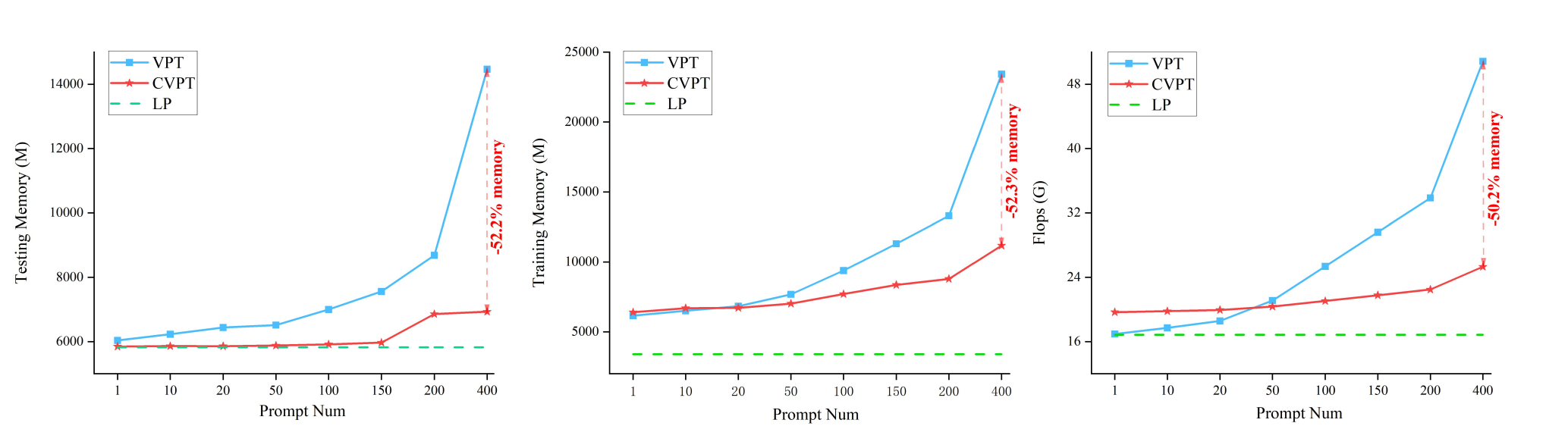} 
    \vspace{-1.5em}
    \caption{
    \textbf{The trends of training memory, testing memory, and Flops with the variation in the number of prompt tokens.} Where LP represents Linear Probing which only tunes the final classifier linear. The batch\_size is set to 32. Pre-trained model is ViT-B/16.}
    \label{fig4}
    \vspace{-1em}
\end{figure*}

\noindent\textbf{Efficiency improvement. } To explore the improvement in the efficiency of CVPT, we also recorded the amount of GPU memory occupied by VPT and CVPT during training and testing as well as the total computation of the two when conducting the above experiments, and the results are shown in Fig.\ref{fig4} follows.

The results reveal that our CVPT has made significant efficiency improvements compared to VPT especially given a large amount of prompt tokens. Additionally, the weight-sharing mechanism allows for targeted optimization in engineering applications, letting cross-attention and self-attention share memory, further widening the efficiency gap with VPT. Moreover, the careful design of CVPT prevents explosive growth in memory and computation as the number of prompts increases. This means we can improve the performance of CVPT by increasing the number of prompts more computationally and efficiently than before. 

In summary, \textbf{our CVPT significantly improves the performance and efficiency of VPT by introducing cross-attention and the weight-sharing mechanism, especially given a larger number of prompts. }Therefore, it allows us to introduce more prompts to the prompt-based method in an efficient manner, thus improving its performance in OOD datasets.

\subsection{Ablation Studies}
\label{sec5.4}
\textbf{The impact of the location of the Cross-Attention (CA). }We conducted experiments with the following five positions to explore the optimal deployment of CA, and the results of the experiments are displayed in Fig.\ref{tab5-5}:

It can be observed that positions near the self-attention (SA) module (1-3) outperform those near the MLP block (4-5), and we attribute this to the more effective adaptation of the rich, contextual features generated by SA. Prior work \cite{repAdapter} also demonstrated that inserting adapters near SA yields better results. Among all options, inserting prompt tokens after SA (position 3) achieves the best performance. Notably, position 2 offers a compelling trade-off, enabling parallel insertion for improved efficiency with only a negligible drop in performance.

\vspace{-0em}
\begin{figure}[H] 
\captionsetup{justification=centering} 
\centering

\begin{minipage}[t]{0.55\columnwidth} 
\vspace{0pt} 
\centering
\includegraphics[width=\linewidth]{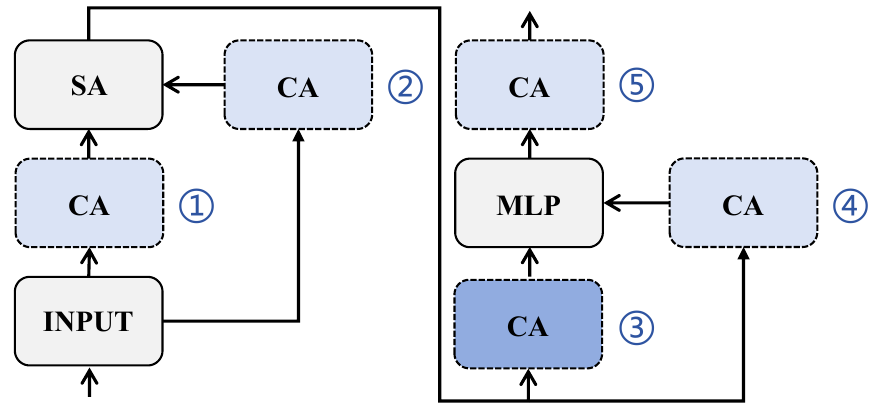} 
\captionsetup{type=figure} 
\end{minipage}
\hfill 
\begin{minipage}[t]{0.4\columnwidth}
\vspace{0pt}
\centering
\captionsetup{type=table} 
\scriptsize 
\begin{tabular}{>{\centering\arraybackslash}p{1.2cm}|>{\centering\arraybackslash}p{1.2cm}}
\toprule
Position & Avg. Acc. \\
\midrule[0.5pt]
1 & 73.9 \\ 
2 & 73.9 \\ 
\rowcolor[RGB]{230,230,230}3 & 74.0 \\ 
4 & 73.3 \\ 
5 & 73.6 \\ 
\bottomrule
\end{tabular}
\end{minipage}
\vspace{-1em}
\caption{(a) The deployments of cross-attention in ViT. Our final deployments are in dark blue. (b) Performance comparisons of different deployments.}
\label{tab5-5} 
\end{figure}

\vspace{-1em}
\noindent\textbf{The impact of weight-sharing between CA and SA. }Weight-sharing can be viewed as an initialization strategy for CA. We introduce random init for comparison and include linear probing (fine-tuning only the classifier head) to demonstrate the effectiveness of introduced prompts and frozen CA. Results with 10 prompts are shown in Table.\ref{tab5-6}:
\begin{table}[h]
\caption{\textbf{Performance comparisons of learnable CA and frozen CA with weight-sharing.}}
\vspace{-1em}
\centering
\scalebox{0.68}{
\begin{tabular}{c|c|c|c|c|c|c}
\hline
Method&Nat.&Spe.&Str.&Avg&FGVC&Param(M)\\
\hline
Linear probing & 68.9 & 77.2 & 26.8 & 57.6 & 79.3&0 \\
Random init+Leanable CA & 80.0 & 84.9 & 57.2 & 74.0 & 89.5&28.4 \\
Weight Sharing+Leanable CA & 79.8 & 85.3 & 58.8 & 74.6 & 89.5&28.4 \\
\hline
Random init+Frozen CA & 72.8 & 81.8 & 36.6 & 63.7 & 86.0&0.09 \\
\rowcolor[RGB]{230, 230, 230}Weight Sharing+Frozen CA & 80.0 & 84.5 & 57.5 & 74.0 & 89.3&0.09 \\
\hline
\end{tabular}}

\label{tab5-6}
\vspace{-0.5em}
\end{table}

After introducing weight-sharing, a frozen  module significantly outperforms one with random initialization. Notably, its performance is comparable to that of a fully learnable CA, despite requiring substantially fewer parameters. This strongly suggests that the parameters inherited from SA provide a powerful inductive bias, effectively guiding the CA module.

While a larger parameter count can increase model capacity, it does not guarantee better performance and may lead to overfitting. The frozen CA, with its smaller and well-initialized parameter space, is therefore easier to optimize and converges much faster than the fully learnable version, proving to be a more efficient approach.

\section{Conclusion}
In the current field of visual fine-tuning, many researchers overlook prompts in favor of adapters due to their strong performance. The few prompt-based derived works do not realize the drawbacks of combining prompts with embedded tokens, continuing to use the method from VPT. In light of this, we thoroughly analyzed the shortcomings of such deployment and proposed CVPT. Its advantages are as follows: 1) It decouples prompts from self-attention and uses cross-attention to establish a connection with embedded tokens. 2) It employs weight-sharing to avoid the large number of learnable parameters introduced by cross-attention. Additionally, we conducted extensive experiments on CVPT, demonstrating its efficiency and performance improvements over VPT and the effectiveness of cross-attention and weight-sharing. Therefore, we prove that prompt-based methods can perform comparably to advanced adapter methods in the visual fine-tuning domain. We hope our work will inspire prompt-based PEFT methods in the future. 

The limitation of our work is that we do not propose new strategies for the init of prompt tokens. We have made some attempts and noticed that other studies have been introduced~\cite{vpt,spt-deep}. However, neither our attempts nor those reported elsewhere have shown substantial effects. Therefore, we follow the same with VPT. We think the exploration of initialization will help us understand how prompts help the model's fine-tuning.
\section{Acknowledgement}
This work was supported in part by the National Natural Science Foundation of China under Grant 62133005, Grant 62293512 and Grant 62293515, the Special funding support for the construction of innovative provinces in Hunan Province under Grant 2021GK1010, the Central Guidance Fund Project for Hunan Science and Technology Development under Grant 2023ZYT003-1.

{
    \small
    \bibliographystyle{ieeenat_fullname}
    \bibliography{main}

\begin{thebibliography}{59}
\providecommand{\natexlab}[1]{#1}
\providecommand{\url}[1]{\texttt{#1}}
\expandafter\ifx\csname urlstyle\endcsname\relax
  \providecommand{\doi}[1]{doi: #1}\else
  \providecommand{\doi}{doi: \begingroup \urlstyle{rm}\Url}\fi

\bibitem[Bahng et~al.(2022)Bahng, Jahanian, Sankaranarayanan, and Isola]{VP}
Hyojin Bahng, Ali Jahanian, Swami Sankaranarayanan, and Phillip Isola.
\newblock Exploring visual prompts for adapting large-scale models.
\newblock 2022.

\bibitem[Bandara and Patel(2024)]{apt}
Wele Gedara~Chaminda Bandara and Vishal~M Patel.
\newblock Attention prompt tuning: Parameter-efficient adaptation of pre-trained models for action recognition.
\newblock In \emph{2024 IEEE 18th International Conference on Automatic Face and Gesture Recognition (FG)}, pages 1--10. IEEE, 2024.

\bibitem[Brown et~al.(2020)Brown, Mann, Ryder, Subbiah, Kaplan, Dhariwal, Neelakantan, Shyam, Sastry, Askell, Agarwal, Herbert{-}Voss, Krueger, Henighan, Child, Ramesh, Ziegler, Wu, Winter, Hesse, Chen, Sigler, Litwin, Gray, Chess, Clark, Berner, McCandlish, Radford, Sutskever, and Amodei]{gpt3}
Tom~B. Brown, Benjamin Mann, Nick Ryder, Melanie Subbiah, Jared Kaplan, Prafulla Dhariwal, Arvind Neelakantan, Pranav Shyam, Girish Sastry, Amanda Askell, Sandhini Agarwal, Ariel Herbert{-}Voss, Gretchen Krueger, Tom Henighan, Rewon Child, Aditya Ramesh, Daniel~M. Ziegler, Jeffrey Wu, Clemens Winter, Christopher Hesse, Mark Chen, Eric Sigler, Mateusz Litwin, Scott Gray, Benjamin Chess, Jack Clark, Christopher Berner, Sam McCandlish, Alec Radford, Ilya Sutskever, and Dario Amodei.
\newblock Language models are few-shot learners.
\newblock \emph{CoRR}, abs/2005.14165, 2020.

\bibitem[Chen et~al.(2021)Chen, Fan, and Panda]{crossvit}
Chun-Fu~Richard Chen, Quanfu Fan, and Rameswar Panda.
\newblock Crossvit: Cross-attention multi-scale vision transformer for image classification.
\newblock In \emph{Proceedings of the IEEE/CVF international conference on computer vision (ICCV)}, pages 357--366, 2021.

\bibitem[Chen et~al.(2022{\natexlab{a}})Chen, Ge, Tong, Wang, Song, Wang, and Luo]{adaptformer}
Shoufa Chen, Chongjian Ge, Zhan Tong, Jiangliu Wang, Yibing Song, Jue Wang, and Ping Luo.
\newblock Adaptformer: Adapting vision transformers for scalable visual recognition.
\newblock \emph{CoRR}, abs/2205.13535, 2022{\natexlab{a}}.

\bibitem[Chen et~al.(2022{\natexlab{b}})Chen, Duan, Wang, He, Lu, Dai, and Qiao]{vit-adapter}
Zhe Chen, Yuchen Duan, Wenhai Wang, Junjun He, Tong Lu, Jifeng Dai, and Yu Qiao.
\newblock Vision transformer adapter for dense predictions.
\newblock \emph{arXiv preprint arXiv:2205.08534}, 2022{\natexlab{b}}.

\bibitem[Clark et~al.(2020)Clark, Luong, Le, and Manning]{clark2020electra}
Kevin Clark, Minh-Thang Luong, Quoc~V Le, and Christopher~D Manning.
\newblock Electra: Pre-training text encoders as discriminators rather than generators.
\newblock \emph{arXiv preprint arXiv:2003.10555}, 2020.

\bibitem[Darcet et~al.(2024)Darcet, Oquab, Mairal, and Bojanowski]{register}
Timoth{\'e}e Darcet, Maxime Oquab, Julien Mairal, and Piotr Bojanowski.
\newblock Vision transformers need registers.
\newblock In \emph{The Twelfth International Conference on Learning Representations (ICLR)}, 2024.

\bibitem[Das et~al.(2023)Das, Dukler, Ravichandran, and Swaminathan]{express}
Rajshekhar Das, Yonatan Dukler, Avinash Ravichandran, and Ashwin Swaminathan.
\newblock Learning expressive prompting with residuals for vision transformers.
\newblock In \emph{2023 IEEE/CVF Conference on Computer Vision and Pattern Recognition (CVPR)}, pages 3366--3377. IEEE Computer Society, 2023.

\bibitem[Derakhshani et~al.(2023)Derakhshani, Sanchez, Bulat, da~Costa, Snoek, Tzimiropoulos, and Martinez]{bayesian-prompt}
Mohammad~Mahdi Derakhshani, Enrique Sanchez, Adrian Bulat, Victor G~Turrisi da Costa, Cees~GM Snoek, Georgios Tzimiropoulos, and Brais Martinez.
\newblock Bayesian prompt learning for image-language model generalization.
\newblock In \emph{Proceedings of the IEEE/CVF International Conference on Computer Vision (ICCV)}, pages 15237--15246, 2023.

\bibitem[Devlin et~al.(2019)Devlin, Chang, Lee, and Toutanova]{bert}
Jacob Devlin, Ming{-}Wei Chang, Kenton Lee, and Kristina Toutanova.
\newblock {BERT:} pre-training of deep bidirectional transformers for language understanding.
\newblock In \emph{NAACL-HLT}, 2019.

\bibitem[Dong et~al.(2024)Dong, Zhang, Chen, Yan, Lin, Yan, Wang, and Yang]{rlrr}
Wei Dong, Xing Zhang, Bihui Chen, Dawei Yan, Zhijun Lin, Qingsen Yan, Peng Wang, and Yang Yang.
\newblock Low-rank rescaled vision transformer fine-tuning: A residual design approach.
\newblock In \emph{Proceedings of the IEEE/CVF Conference on Computer Vision and Pattern Recognition (CVPR)}, pages 16101--16110, 2024.

\bibitem[Fang et~al.(2024)Fang, Wang, Yi, and Ma]{dmlora}
Zhengyi Fang, Yue Wang, Ran Yi, and Lizhuang Ma.
\newblock Dropout mixture low-rank adaptation for visual parameters-efficient fine-tuning.
\newblock In \emph{European Conference on Computer Vision (ECCV)}, pages 369--386. Springer, 2024.

\bibitem[Fei-Fei et~al.(2006)Fei-Fei, Fergus, and Perona]{caltech}
Li Fei-Fei, R. Fergus, and P. Perona.
\newblock One-shot learning of object categories.
\newblock \emph{IEEE Transactions on Pattern Analysis and Machine Intelligence}, page 594–611, 2006.

\bibitem[Gebru et~al.(2017)Gebru, Krause, Wang, Chen, Deng, and Fei-Fei]{cars}
Timnit Gebru, Jonathan Krause, Yilun Wang, Duyun Chen, Jia Deng, and Li Fei-Fei.
\newblock Fine-grained car detection for visual census estimation.
\newblock \emph{Proceedings of the AAAI conference on artificial intelligence (AAAI)}, 2017.

\bibitem[Geiger et~al.(2013)Geiger, Lenz, Stiller, and Urtasun]{kitti}
A Geiger, P Lenz, C Stiller, and R Urtasun.
\newblock Vision meets robotics: The kitti dataset.
\newblock \emph{The International Journal of Robotics Research}, page 1231–1237, 2013.

\bibitem[Han et~al.()Han, Wang, Cui, Wang, Huang, Qi, and Liu]{facing}
Cheng Han, Qifan Wang, Yiming Cui, Wenguan Wang, Lifu Huang, Siyuan Qi, and Dongfang Liu.
\newblock Facing the elephant in the room: Visual prompt tuning or full finetuning?
\newblock In \emph{The Twelfth International Conference on Learning Representations (ICLR)}.

\bibitem[Han et~al.(2023)Han, Wang, Cui, Cao, Wang, Qi, and Liu]{e2vpt}
Cheng Han, Qifan Wang, Yiming Cui, Zhiwen Cao, Wenguan Wang, Siyuan Qi, and Dongfang Liu.
\newblock E2vpt: An effective and efficient approach for visual prompt tuning.
\newblock In \emph{2023 IEEE/CVF International Conference on Computer Vision (ICCV)}, pages 17445--17456. IEEE Computer Society, 2023.

\bibitem[He et~al.(2023)He, Cai, Zhang, Tao, and Zhuang]{spt}
Haoyu He, Jianfei Cai, Jing Zhang, Dacheng Tao, and Bohan Zhuang.
\newblock Sensitivity-aware visual parameter-efficient fine-tuning.
\newblock In \emph{2023 IEEE/CVF International Conference on Computer Vision (ICCV)}, pages 11791--11801. IEEE Computer Society, 2023.

\bibitem[He et~al.(2016)He, Zhang, Ren, and Sun]{resnet}
Kaiming He, Xiangyu Zhang, Shaoqing Ren, and Jian Sun.
\newblock Deep residual learning for image recognition.
\newblock In \emph{CVPR}, 2016.

\bibitem[He et~al.(2022)He, Chen, Xie, Li, Doll{\'{a}}r, and Girshick]{mae}
Kaiming He, Xinlei Chen, Saining Xie, Yanghao Li, Piotr Doll{\'{a}}r, and Ross~B. Girshick.
\newblock Masked autoencoders are scalable vision learners.
\newblock In \emph{CVPR}, 2022.

\bibitem[He et~al.(2020)He, Liu, Gao, and Chen]{deberta}
Pengcheng He, Xiaodong Liu, Jianfeng Gao, and Weizhu Chen.
\newblock Deberta: Decoding-enhanced bert with disentangled attention.
\newblock \emph{arXiv preprint arXiv:2006.03654}, 2020.

\bibitem[Helber et~al.(2019)Helber, Bischke, Dengel, and Borth]{erousat}
Patrick Helber, Benjamin Bischke, Andreas Dengel, and Damian Borth.
\newblock Eurosat: A novel dataset and deep learning benchmark for land use and land cover classification.
\newblock \emph{IEEE Journal of Selected Topics in Applied Earth Observations and Remote Sensing}, page 2217–2226, 2019.

\bibitem[Houlsby et~al.(2019)Houlsby, Giurgiu, Jastrzebski, Morrone, de~Laroussilhe, Gesmundo, Attariyan, and Gelly]{adapter}
Neil Houlsby, Andrei Giurgiu, Stanislaw Jastrzebski, Bruna Morrone, Quentin de Laroussilhe, Andrea Gesmundo, Mona Attariyan, and Sylvain Gelly.
\newblock Parameter-efficient transfer learning for {NLP}.
\newblock In \emph{ICML}, 2019.

\bibitem[Hu et~al.(2022)Hu, yelong shen, Wallis, Allen-Zhu, Li, Wang, Wang, and Chen]{lora}
Edward~J Hu, yelong shen, Phillip Wallis, Zeyuan Allen-Zhu, Yuanzhi Li, Shean Wang, Lu Wang, and Weizhu Chen.
\newblock Lo{RA}: Low-rank adaptation of large language models.
\newblock In \emph{ICLR}, 2022.

\bibitem[Huang et~al.(2023)Huang, Dong, Chen, Zhang, Wang, Hua, and Yu]{dam}
Qidong Huang, Xiaoyi Dong, Dongdong Chen, Weiming Zhang, Feifei Wang, Gang Hua, and Nenghai Yu.
\newblock Diversity-aware meta visual prompting.
\newblock In \emph{2023 IEEE/CVF Conference on Computer Vision and Pattern Recognition (CVPR)}, pages 10878--10887. IEEE Computer Society, 2023.

\bibitem[Jia et~al.(2022)Jia, Tang, Chen, Cardie, Belongie, Hariharan, and Lim]{vpt}
Menglin Jia, Luming Tang, Bor-Chun Chen, Claire Cardie, Serge Belongie, Bharath Hariharan, and Ser-Nam Lim.
\newblock Visual prompt tuning.
\newblock In \emph{European Conference on Computer Vision (ECCV)}, pages 709--727. Springer, 2022.

\bibitem[Johnson et~al.(2017)Johnson, Hariharan, van~der Maaten, Fei-Fei, Zitnick, and Girshick]{clevr}
Justin Johnson, Bharath Hariharan, Laurens van~der Maaten, Li Fei-Fei, C.~Lawrence Zitnick, and Ross Girshick.
\newblock Clevr: A diagnostic dataset for compositional language and elementary visual reasoning.
\newblock In \emph{2017 IEEE Conference on Computer Vision and Pattern Recognition (CVPR)}, 2017.

\bibitem[Khosla et~al.()Khosla, Jayadevaprakash, Yao, and Li]{dogs}
Aditya Khosla, Nityananda Jayadevaprakash, Bangpeng Yao, and Fei-Fei Li.
\newblock Novel dataset for fine-grained image categorization: Stanford dogs.

\bibitem[Krizhevsky(2009)]{cifar}
Alex Krizhevsky.
\newblock Learning multiple layers of features from tiny images.
\newblock 2009.

\bibitem[Krizhevsky et~al.(2012)Krizhevsky, Sutskever, and Hinton]{alex}
Alex Krizhevsky, Ilya Sutskever, and Geoffrey~E Hinton.
\newblock Imagenet classification with deep convolutional neural networks.
\newblock In \emph{NIPS}, 2012.

\bibitem[Le et~al.(2025)Le, Nguyen, Nguyen, Nguyen, and Ho]{vapt}
Minh Le, Anh Nguyen, Huy Nguyen, Chau Nguyen, and Nhat Ho.
\newblock Adaptive prompt: Unlocking the power of visual prompt tuning.
\newblock \emph{arXiv preprint arXiv:2501.18936}, 2025.

\bibitem[Lester et~al.(2021)Lester, Al-Rfou, and Constant]{prompt-tuning}
Brian Lester, Rami Al-Rfou, and Noah Constant.
\newblock The power of scale for parameter-efficient prompt tuning.
\newblock \emph{arXiv preprint arXiv:2104.08691}, 2021.

\bibitem[Lewis et~al.(2019)Lewis, Liu, Goyal, Ghazvininejad, Mohamed, Levy, Stoyanov, and Zettlemoyer]{bart}
Mike Lewis, Yinhan Liu, Naman Goyal, Marjan Ghazvininejad, Abdelrahman Mohamed, Omer Levy, Ves Stoyanov, and Luke Zettlemoyer.
\newblock Bart: Denoising sequence-to-sequence pre-training for natural language generation, translation, and comprehension.
\newblock \emph{arXiv preprint arXiv:1910.13461}, 2019.

\bibitem[Liu et~al.(2019)Liu, Ott, Goyal, Du, Joshi, Chen, Levy, Lewis, Zettlemoyer, and Stoyanov]{Roberta}
Yinhan Liu, Myle Ott, Naman Goyal, Jingfei Du, Mandar Joshi, Danqi Chen, Omer Levy, Mike Lewis, Luke Zettlemoyer, and Veselin Stoyanov.
\newblock Roberta: A robustly optimized bert pretraining approach.
\newblock \emph{arXiv preprint arXiv:1907.11692}, 2019.

\bibitem[Luo et~al.(2023)Luo, Huang, Zhou, Sun, Jiang, Wang, and Ji]{repAdapter}
Gen Luo, Minglang Huang, Yiyi Zhou, Xiaoshuai Sun, Guannan Jiang, Zhiyu Wang, and Rongrong Ji.
\newblock Towards efficient visual adaption via structural re-parameterization.
\newblock \emph{arXiv preprint arXiv:2302.08106}, 2023.

\bibitem[Matthey et~al.(2017)Matthey, Higgins, Hassabis, and Lerchner]{matthey2017dsprites}
Loic Matthey, Irina Higgins, Demis Hassabis, and Alexander Lerchner.
\newblock dsprites: Disentanglement testing sprites dataset, 2017.

\bibitem[Netzer et~al.(2011)Netzer, Wang, Coates, Bissacco, Wu, and Ng]{svhn}
Yuval Netzer, Tao Wang, Adam Coates, Alessandro Bissacco, Bo Wu, and AndrewY. Ng.
\newblock Reading digits in natural images with unsupervised feature learning.
\newblock 2011.

\bibitem[Nilsback and Zisserman(2006)]{flower}
M-E Nilsback and Andrew Zisserman.
\newblock A visual vocabulary for flower classification.
\newblock In \emph{CVPR}, 2006.

\bibitem[Parkhi et~al.(2012)Parkhi, Vedaldi, Zisserman, and Jawahar]{pets}
Omkar~M Parkhi, Andrea Vedaldi, Andrew Zisserman, and CV Jawahar.
\newblock Cats and dogs.
\newblock In \emph{CVPR}, 2012.

\bibitem[Pei et~al.(2024)Pei, Xia, Chen, Li, Tian, and Lu]{sa2vp}
Wenjie Pei, Tongqi Xia, Fanglin Chen, Jinsong Li, Jiandong Tian, and Guangming Lu.
\newblock Sa$^2$vp: Spatially aligned-and-adapted visual prompt.
\newblock In \emph{Proceedings of the AAAI conference on artificial intelligence (AAAI)}, pages 4450--4458, 2024.

\bibitem[Radford et~al.(2018)Radford, Narasimhan, Salimans, Sutskever, et~al.]{gpt-1}
Alec Radford, Karthik Narasimhan, Tim Salimans, Ilya Sutskever, et~al.
\newblock Improving language understanding by generative pre-training.
\newblock 2018.

\bibitem[Radford et~al.(2019)Radford, Wu, Child, Luan, Amodei, Sutskever, et~al.]{gpt-2}
Alec Radford, Jeffrey Wu, Rewon Child, David Luan, Dario Amodei, Ilya Sutskever, et~al.
\newblock Language models are unsupervised multitask learners.
\newblock \emph{OpenAI blog}, 1\penalty0 (8):\penalty0 9, 2019.

\bibitem[Radford et~al.(2021)Radford, Kim, Hallacy, Ramesh, Goh, Agarwal, Sastry, Askell, Mishkin, Clark, Krueger, and Sutskever]{clip}
Alec Radford, Jong~Wook Kim, Chris Hallacy, Aditya Ramesh, Gabriel Goh, Sandhini Agarwal, Girish Sastry, Amanda Askell, Pamela Mishkin, Jack Clark, Gretchen Krueger, and Ilya Sutskever.
\newblock Learning transferable visual models from natural language supervision.
\newblock In \emph{ICML}, 2021.

\bibitem[Sung et~al.(2022)Sung, Cho, and Bansal]{vl-adapter}
Yi-Lin Sung, Jaemin Cho, and Mohit Bansal.
\newblock Vl-adapter: Parameter-efficient transfer learning for vision-and-language tasks.
\newblock In \emph{Proceedings of the IEEE/CVF Conference on Computer Vision and Pattern Recognition}, pages 5227--5237, 2022.

\bibitem[Van~Horn et~al.(2015)Van~Horn, Branson, Farrell, Haber, Barry, Ipeirotis, Perona, and Belongie]{birds}
Grant Van~Horn, Steve Branson, Ryan Farrell, Scott Haber, Jessie Barry, Panos Ipeirotis, Pietro Perona, and Serge Belongie.
\newblock Building a bird recognition app and large scale dataset with citizen scientists: The fine print in fine-grained dataset collection.
\newblock In \emph{2015 IEEE Conference on Computer Vision and Pattern Recognition (CVPR)}, 2015.

\bibitem[Vaswani et~al.(2017)Vaswani, Shazeer, Parmar, Uszkoreit, Jones, Gomez, Kaiser, and Polosukhin]{vaswani2017attention}
Ashish Vaswani, Noam Shazeer, Niki Parmar, Jakob Uszkoreit, Llion Jones, Aidan~N Gomez, {\L}ukasz Kaiser, and Illia Polosukhin.
\newblock Attention is all you need.
\newblock \emph{Advances in neural information processing systems (NeuIPS)}, 30, 2017.

\bibitem[Veeling et~al.(2018)Veeling, Linmans, Winkens, Cohen, and Welling]{patchcamelyon}
Bastiaan~S. Veeling, Jasper Linmans, Jim Winkens, Taco Cohen, and Max Welling.
\newblock \emph{Rotation Equivariant CNNs for Digital Pathology}, page 210–218.
\newblock 2018.

\bibitem[Wah et~al.(2011)Wah, Branson, Welinder, Perona, and Belongie]{CUB}
Catherine Wah, Steve Branson, Peter Welinder, Pietro Perona, and Serge Belongie.
\newblock The caltech-ucsd birds-200-2011 dataset.
\newblock 2011.

\bibitem[Wang et~al.(2024)Wang, Cheng, Fang, Zhang, Duan, and Wang]{spt-deep}
Yuzhu Wang, Lechao Cheng, Chaowei Fang, Dingwen Zhang, Manni Duan, and Meng Wang.
\newblock Revisiting the power of prompt for visual tuning.
\newblock In \emph{International Conference on Machine Learning (ICLR)}, pages 50233--50247. PMLR, 2024.

\bibitem[Xiao et~al.(2024)Xiao, Tian, Chen, Han, and Lewis]{attentionsink}
Guangxuan Xiao, Yuandong Tian, Beidi Chen, Song Han, and Mike Lewis.
\newblock Efficient streaming language models with attention sinks.
\newblock In \emph{The Twelfth International Conference on Learning Representations (ICLR)}, 2024.

\bibitem[Xie et~al.(2017)Xie, Girshick, Doll{\'a}r, Tu, and He]{resnext}
Saining Xie, Ross Girshick, Piotr Doll{\'a}r, Zhuowen Tu, and Kaiming He.
\newblock Aggregated residual transformations for deep neural networks.
\newblock In \emph{Proceedings of the IEEE conference on computer vision and pattern recognition}, pages 1492--1500, 2017.

\bibitem[Yang et~al.(2023)Yang, Zhu, Xie, Zhang, Chen, and Li]{aim}
Taojiannan Yang, Yi Zhu, Yusheng Xie, Aston Zhang, Chen Chen, and Mu Li.
\newblock Aim: Adapting image models for efficient video action recognition.
\newblock \emph{arXiv preprint arXiv:2302.03024}, 2023.

\bibitem[Yang et~al.(2019)Yang, Dai, Yang, Carbonell, Salakhutdinov, and Le]{xlnet}
Zhilin Yang, Zihang Dai, Yiming Yang, Jaime Carbonell, Russ~R Salakhutdinov, and Quoc~V Le.
\newblock Xlnet: Generalized autoregressive pretraining for language understanding.
\newblock \emph{Advances in neural information processing systems (NeurIPS)}, 32, 2019.

\bibitem[Zaken et~al.(2022)Zaken, Goldberg, and Ravfogel]{bitfit}
Elad~Ben Zaken, Yoav Goldberg, and Shauli Ravfogel.
\newblock Bitfit: Simple parameter-efficient fine-tuning for transformer-based masked language-models.
\newblock In \emph{ACL}, 2022.

\bibitem[Zeng et~al.(2025)Zeng, Han, Wang, Wu, Geng, Huangg, Wu, and Liu]{vfpt}
Runjia Zeng, Cheng Han, Qifan Wang, Chunshu Wu, Tong Geng, Lifu Huangg, Ying~Nian Wu, and Dongfang Liu.
\newblock Visual fourier prompt tuning.
\newblock \emph{Advances in Neural Information Processing Systems (NeurIPS)}, 37:\penalty0 5552--5585, 2025.

\bibitem[Zhai et~al.(2022)Zhai, Kolesnikov, Houlsby, and Beyer]{vitg}
Xiaohua Zhai, Alexander Kolesnikov, Neil Houlsby, and Lucas Beyer.
\newblock Scaling vision transformers.
\newblock In \emph{CVPR}, 2022.

\bibitem[Zheng et~al.(2021)Zheng, Lu, Zhao, Zhu, Luo, Wang, Fu, Feng, Xiang, Torr, et~al.]{setr}
Sixiao Zheng, Jiachen Lu, Hengshuang Zhao, Xiatian Zhu, Zekun Luo, Yabiao Wang, Yanwei Fu, Jianfeng Feng, Tao Xiang, Philip~HS Torr, et~al.
\newblock Rethinking semantic segmentation from a sequence-to-sequence perspective with transformers.
\newblock In \emph{Proceedings of the IEEE/CVF conference on computer vision and pattern recognition}, pages 6881--6890, 2021.

\bibitem[Zhou et~al.(2017)Zhou, Zhao, Puig, Fidler, Barriuso, and Torralba]{ade20k}
Bolei Zhou, Hang Zhao, Xavier Puig, Sanja Fidler, Adela Barriuso, and Antonio Torralba.
\newblock Scene parsing through ade20k dataset.
\newblock In \emph{Proceedings of the IEEE Conference on Computer Vision and Pattern Recognition (CVPR)}, 2017.

\end{thebibliography}
}


\end{document}